\documentclass[letterpaper, 10 pt, conference]{ieeeconf}  %
\pdfoutput=1

\IEEEoverridecommandlockouts                              %

\usepackage{cite}
\usepackage{xstring}
\usepackage{graphicx}
\usepackage{pgfplots}
\usepackage{pgfplotstable}
\usepackage{ifthen}
\usepackage{tikz}
\usepackage{tikz-3dplot}
\usepackage{subfig}
\usepackage{booktabs}
\usepackage{multirow}
\usepackage{url}
\usepackage{amsmath}
\usepackage{amssymb}
\usepackage{pifont}
\usepackage{tikz}
\usepackage{xcolor}
\usepackage{framed}
\usetikzlibrary{shapes,arrows,shadows,positioning,math,backgrounds,calc,intersections}
\pgfplotsset{compat=1.16}
\newcommand{\cmark}{\ding{51}}%

\newcommand{\normal}{\mathcal{N}}
\newcommand{\nef}{\Omega}
\newcommand{\features}{\mathcal{F}}
\newcommand{\head}{\mathcal{H}}
\newcommand{\embedding}{\mathcal{E}}

\newcommand{\loss}{\mathcal{L}}
\newcommand{\real}{\mathbb{R}}
\newcommand{\nat}{\mathbb{N}}

\newcommand{\pose}{\mathcal{P}}
\newcommand{\se}[1]{\mathfrak{se}#1}

\newcommand{\transform}{\mathcal{T}}

\newcommand{\etAl}{\emph{et al.}}

\newcommand{\refFigure}[1]{Fig. \ref{#1}}

\newcommand{\refSection}[1]{section \ref{#1}}

\newcommand{\refTable}[1]{Table \ref{#1}}

\def\goldenratio{1.61}
\def\plothalfcolumnwidth{0.45}

\definecolor{cblue}{HTML}{1f77b4}
\definecolor{corange}{HTML}{ff7f0e}
\definecolor{cgreen}{HTML}{2ca02c}
\definecolor{cred}{HTML}{d62728}
\definecolor{cpurple}{HTML}{9467bd}
\definecolor{cbrown}{HTML}{8c564b}
\definecolor{cpink}{HTML}{e377c2}
\definecolor{cgrey}{HTML}{7f7f7f}
\definecolor{cyellow}{HTML}{bcbd22}
\definecolor{ccyan}{HTML}{17becf}

\pgfplotscreateplotcyclelist{tab10}{
    cblue,mark=*,mark size=1.5, mark options={solid}\\
    densely dashed, corange,mark=*,mark size=1.5, mark options={solid}\\
    densely dotted, cgreen,mark=*,mark size=1.5, mark options={solid}\\
    densely dashdotted, cred,mark=*,mark size=1.5, mark options={solid}\\
    cpurple,mark=*,mark size=1.5, mark options={solid}\\
    densely dashed, cbrown,mark=*,mark size=1.5, mark options={solid}\\
    densely dotted, cpink,mark=*,mark size=1.5, mark options={solid}\\
    densely dashdotted, cgrey,mark=*,mark size=1.5, mark options={solid}\\
    cyellow,mark=*,mark size=1.5, mark options={solid}\\
    ccyan,mark=*,mark size=1.5, mark options={solid}\\
}

\pgfplotscreateplotcyclelist{tab10flat}{
    cblue\\
    densely dashed, corange\\
    densely dotted, cgreen\\
    densely dashdotted, cred\\
    cpurple\\
    densely dashed, cbrown\\
    densely dotted, cpink\\
    densely dashdotted, cgrey\\
    cyellow\\
    ccyan\\
}

\pgfplotsset{
    PlotStyleBase/.style={
        cycle list name=tab10,
        draw=white!80!black
    },
}
\pgfplotsset{
    PlotStyleColumn/.style={
        PlotStyleBase,
        width=\columnwidth,
        height=\columnwidth/\goldenratio,
        legend style={font=\footnotesize},
        tick label style={font=\footnotesize},
        label style={font=\small},
        title style={font=\small},
        every axis title shift=0pt,
        max space between ticks=12,
        every mark/.append style={mark size=6},
        major tick length=0.1cm,
        minor tick length=0.066cm,
    },
}
\pgfplotsset{
    PlotStyleHalfColumn/.style={
        PlotStyleBase,
        width=\plothalfcolumnwidth\columnwidth,
        height=\plothalfcolumnwidth\columnwidth/\goldenratio,
        scale only axis,
        legend style={font=\scriptsize},
        tick label style={font=\scriptsize},
        label style={font=\footnotesize},
        title style={font=\small},
        every axis title shift=0pt,
        max space between ticks=12,
        every mark/.append style={mark size=6},
        major tick length=0.1cm,
        minor tick length=0.066cm,
        every legend image post/.append style={scale=0.8},
    },
}

\title{\LARGE \bf
Neural Semantic Map-Learning for Autonomous Vehicles
}

\author{Markus Herb$^{1,2}$, Nassir Navab$^{1}$ and Federico Tombari$^{1,3}$%
\thanks{$^{1}$Technical University of Munich, Computer-Aided Medical Procedures, Garching bei München, Germany}%
\thanks{$^{2}$CARIAD SE, Ingolstadt, Germany}%
\thanks{$^{3}$Google, Zurich, Switzerland}%
}%

\begin{document}%
\maketitle%
\thispagestyle{empty}%
\pagestyle{empty}%
\tikz[remember picture,overlay] {\node at (current page.south) {\raisebox{2cm}{\parbox{\textwidth}{\footnotesize\centering\copyright~2024 IEEE. Personal use of this material is permitted. Permission from IEEE must be obtained for all other uses, in any current or future media, including reprinting/republishing this material for advertising or promotional purposes, creating new collective works, for resale or redistribution to servers or lists, or reuse of any copyrighted component of this work in other works.}}};}%
\begin{abstract}%
Autonomous vehicles demand detailed maps to maneuver reliably through traffic, which need to be kept up-to-date to ensure a safe operation.
A promising way to adapt the maps to the ever-changing road-network is to use crowd-sourced data from a fleet of vehicles.
In this work, we present a mapping system that fuses local submaps gathered from a fleet of vehicles at a central instance to produce a coherent map of the road environment including drivable area, lane markings, poles, obstacles and more as a 3D mesh.
Each vehicle contributes locally reconstructed submaps as lightweight meshes, making our method applicable to a wide range of reconstruction methods and sensor modalities.
Our method jointly aligns and merges the noisy and incomplete local submaps using a scene-specific Neural Signed Distance Field, which is supervised using the submap meshes to predict a fused environment representation.
We leverage memory-efficient sparse feature-grids to scale to large areas and introduce a confidence score to model uncertainty in scene reconstruction.
Our approach is evaluated on two datasets with different local mapping methods, showing improved pose alignment and reconstruction over existing methods.
Additionally, we demonstrate the benefit of multi-session mapping and examine the required amount of data to enable high-fidelity map learning for autonomous vehicles.
\end{abstract}%
\section{Introduction}

Maps are crucial for many applications such as mobile robots including autonomous vehicles or augmented reality to facilitate localization, navigation and interaction with the environment.
While maps can prove beneficial for higher-level tasks, it is paramount that the map content is accurate and up-to-date, especially in safety critical settings such as autonomous driving.
Traditionally, dedicated mapping vehicles have been used for map data collection, which however quickly becomes infeasible to cover areas beyond city-scale with a high update rate due to the limited number of vehicles and large operating effort.
Collecting crowd-sourced user data from car fleets is a promising alternative to gather fresh mapping data in a timely manner
\cite{Dabeer2017, Pannen2019b, Herb2019iros, Kim2021a}. In contrast to specialized mapping vehicles, commodity cars typically come with a relatively limited sensor set, storage and upload capacities, which demands efficient use of resources to be worthwhile for mapping large scale road networks.

\begin{figure}[tp]%
    \centering%
    \resizebox{\columnwidth}{!}{%
        \input{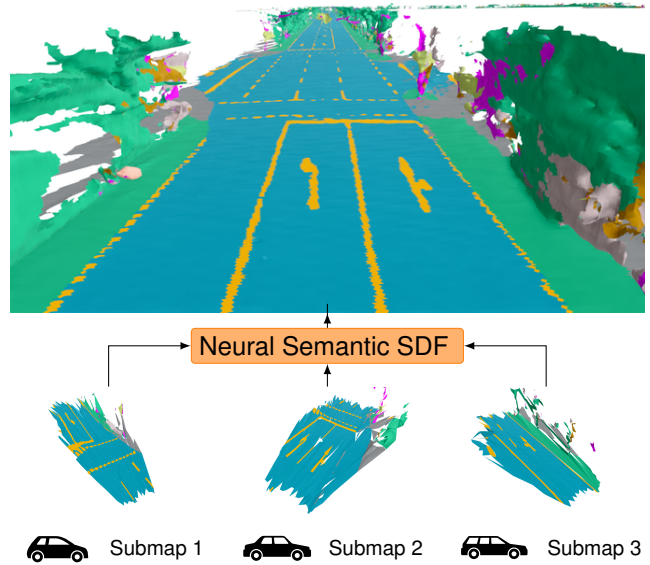}%
    }%
    \caption{Our neural multi-session mapping approach fuses dense submaps created from different agents through local dense SLAM by training a neural field to predict dense semantics and geometry that best explains the submaps collected from multiple mapping sessions.}%
    \label{fig:cloudmap:teaser}%
\end{figure}

To address these challenges, we present a centralized mapping system producing dense semantic maps, depicted in \refFigure{fig:cloudmap:teaser}, by incorporating map contributions from many individual vehicles or \emph{agents}.
We chose to represent the individual submaps as 3D semantic surface meshes, as these can represent dense geometry and semantics without the need for spatial discretization, are lightweight in terms of storage demands and can be produced using various mapping approaches relying on different sensor configurations.
Furthermore, the meshes can often be easily extracted from the local environment representation each agent builds for other tasks such as obstacle avoidance, thus saving computational effort by fusing the local sensor data only once.
Our proposed system hence avoids the need for raw sensor data to be sent to a central server altogether and instead uses already fused submaps as intermediate mapping data uploaded by the agents to the server.
While being friendly for low-bandwidth mobile connections, this also reduces privacy concerns associated with image data.

Our method builds on recent advances in Neural Implicit Representations and uses a Neural Signed Distance Field \cite{Park2019} to represent geometry and semantics of the environment.
During reconstruction, we supervise the neural field to produce a signed distance and semantic logits for each point in space that best matches all the input submaps observed by different agents exploring the scene.
Since the local maps produced by each vehicle are only coarsely aligned using inaccurate GPS information, we also optimize the relative poses of all submaps jointly during the reconstruction.
Our fundamental assumption is that neural fields are capable to learn a \emph{fused, coherent} scene representation from noisy, incomplete and potentially contradicting input submaps observed from different agents.
To enable fast update times and scalability to large regions, we cluster the mapped areas into independent tiles, each represented by a feature OcTree \cite{Takikawa2021} or HashGrid \cite{Mueller2022instantngp} as efficient spatial grid-structure with prediction heads for semantics and geometry.

To summarize, our main contributions in this work are
\begin{itemize}
    \item a novel multi-session mapping system for dense semantic submap fusion using neural fields,
    \item extensive experiments on two real-world datasets demonstrating the efficacy of our method for crowd-sourced map learning
\end{itemize}

\begin{figure*}[t!]
    \centering%
    \resizebox{\textwidth}{!}{%
\tikzstyle{mlp_node}=[draw, thick,circle,minimum size=10,inner sep=0.5,outer sep=0.6]%
\tikzstyle{mlp_connect}=[thick]%
\begin{tikzpicture}[
    arrowstyle/.style={->,shorten >=2pt,shorten <=2pt,>=stealth},
    revarrowstyle/.style={<-,shorten >=2pt,shorten <=2pt,>=stealth},
    arrowstyle_train/.style={->,shorten >=2pt,shorten <=2pt,>=stealth},
    arrowstyle_infer/.style={dotted,->,shorten >=2pt,shorten <=2pt,>=stealth},
    arrowstyle_tight/.style={->,>=stealth},
    linestyle_tight/.style={-},
    processstyle/.style={draw=cblue, fill=cblue!60!white, rounded corners=.05cm,inner sep=3pt, align=center},
    posestyle/.style={draw=cred, fill=cred!60!white, rounded corners=.05cm,inner sep=3pt, align=center},
    nnblockstyle/.style={draw=cgreen, fill=cgreen!60!white, rounded corners=.05cm,inner sep=3pt, align=center},
    nnstatestyle/.style={draw, fill=white, rounded corners=.05cm,inner sep=3pt, align=center},
    headstyle/.style={draw=corange, fill=corange!60!white, rounded corners=.05cm,inner sep=3pt, align=center},
    fieldstyle/.style={draw=ccyan, fill=ccyan!60!white, rounded corners=.025cm,inner sep=1pt, align=center}
]

\draw [fill=none,dashed,cgrey,thick] (-3.75,1.5) rectangle (-0.35,-3);
\draw [fill=none,dashed,cgrey,thick] (-0.1,1.5) rectangle (18,-3);

\node[inner sep=0pt, cgrey] at (-2.75,1.25) (local_map_label)  {
    \footnotesize\sf Local Mapping
};
\node[inner sep=0pt, cgrey] at (15.5,1.25) (fusion_map_label)  {
    \footnotesize\sf Server-based Multi-Session Fusion
};

\node[inner sep=0pt] at (-1.5, 0.15) (map1)  {
    \includegraphics[width=1.5cm]{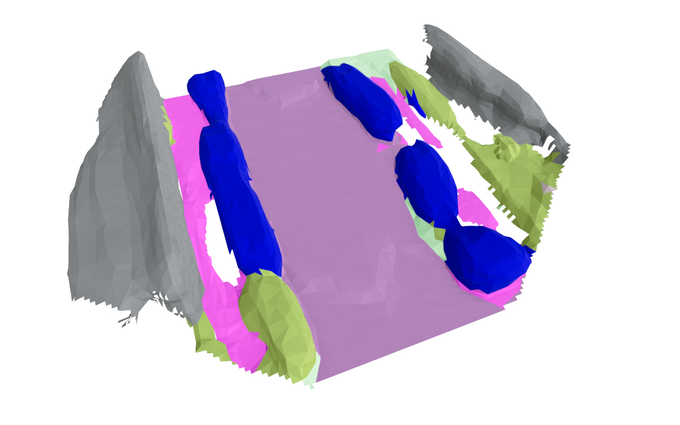}
};
\node[inner sep=0pt, below=2mm of map1] (map2)  {
    \includegraphics[width=1.5cm]{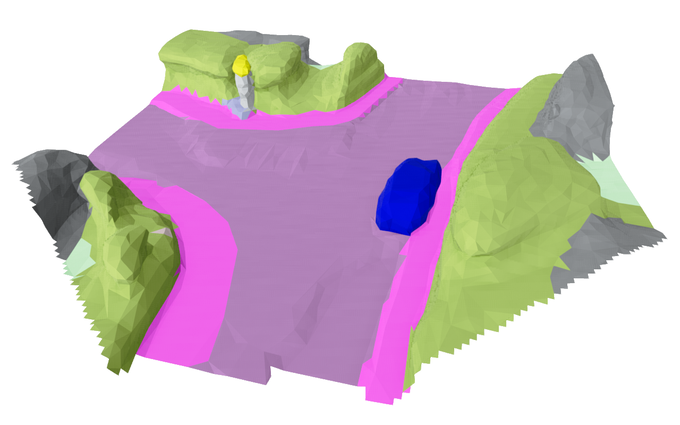}
};
\node[inner sep=0pt, below=2mm of map2] (map3)  {
    \includegraphics[width=1.5cm]{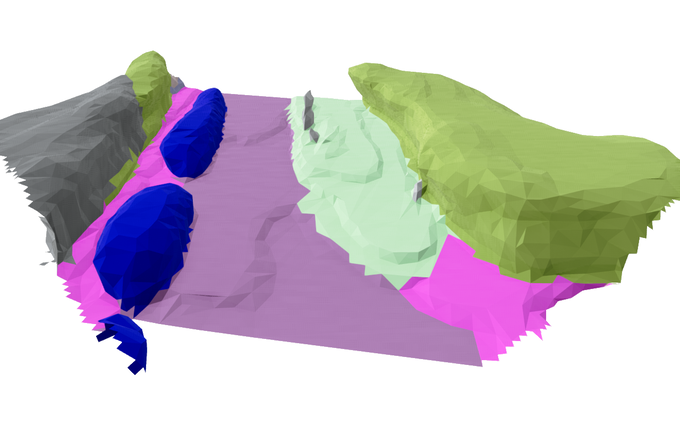}
};
\node[inner sep=0pt, above=1mm of map1] (map_label)  {
    \footnotesize\sf Submaps
};

\node[inner sep=0pt, left=2mm of map1] (car1)  {
    \includegraphics[width=1cm,trim=2cm 2cm 2cm 2cm,clip]{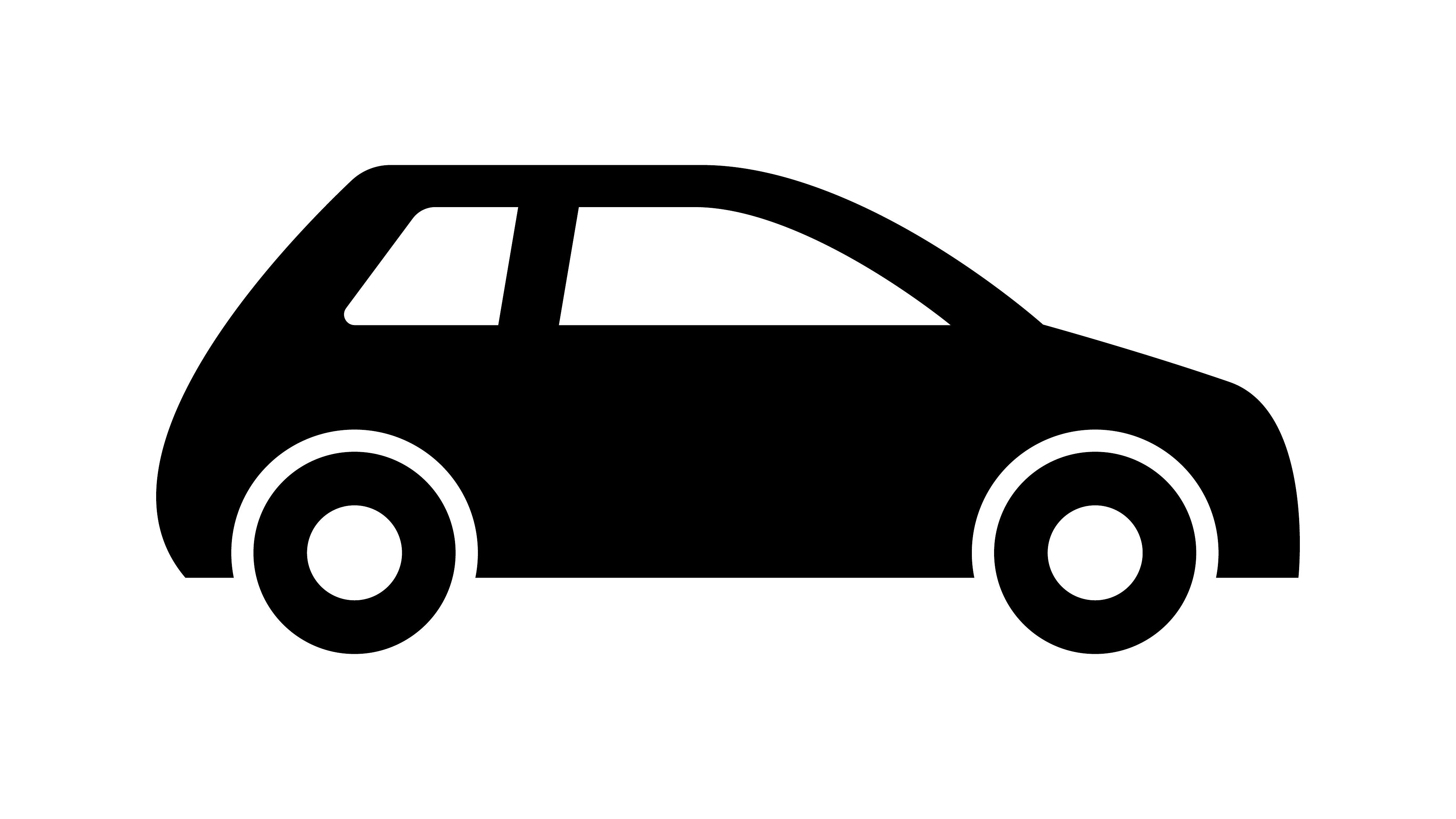}
};
\node[inner sep=0pt, left=2mm of map2] (car2)  {
    \includegraphics[width=1cm,trim=2cm 2cm 2cm 2cm,clip]{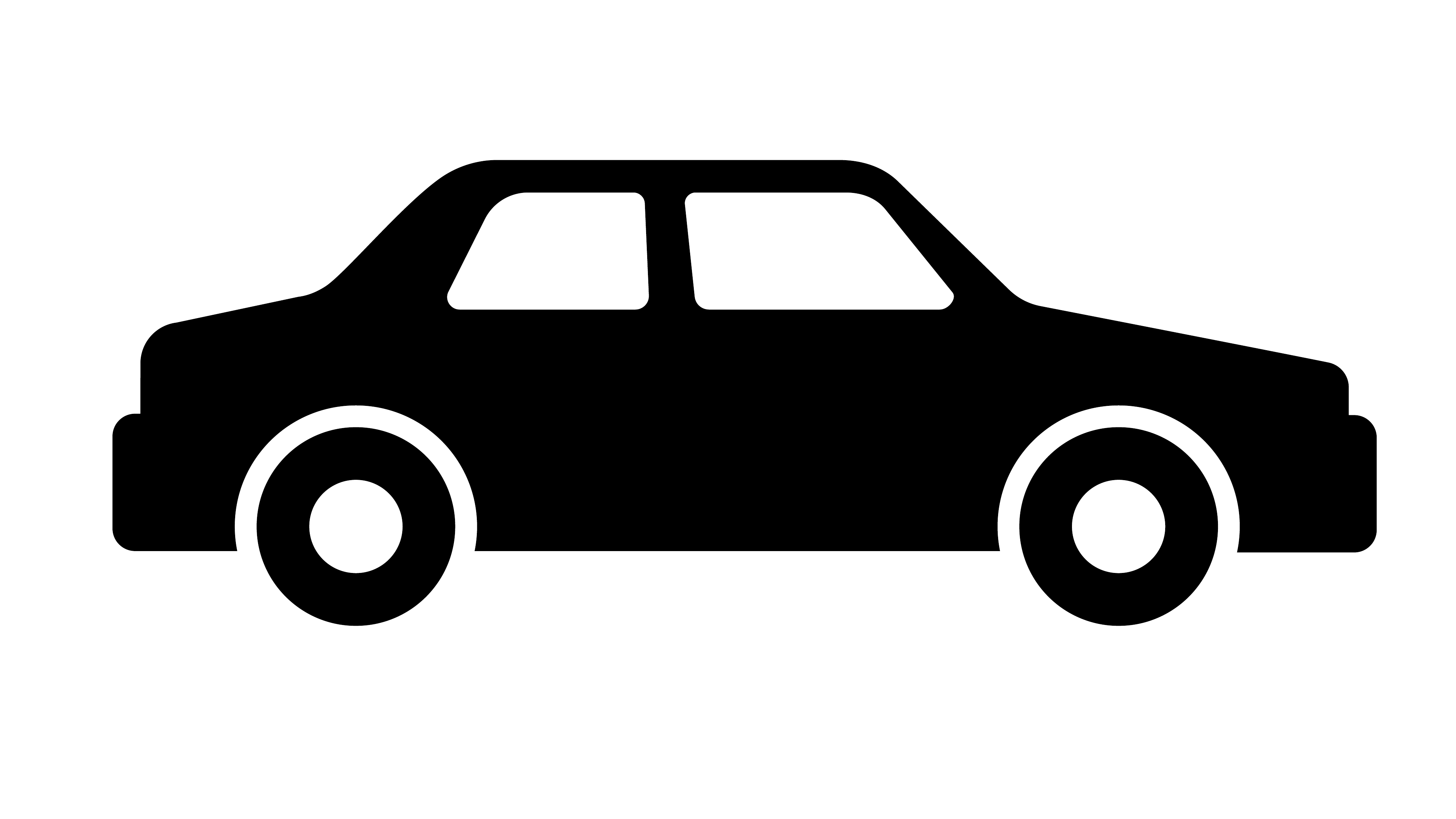}
};
\node[inner sep=0pt, left=2mm of map3] (car3)  {
    \includegraphics[width=1cm,trim=2cm 2cm 2cm 2cm,clip]{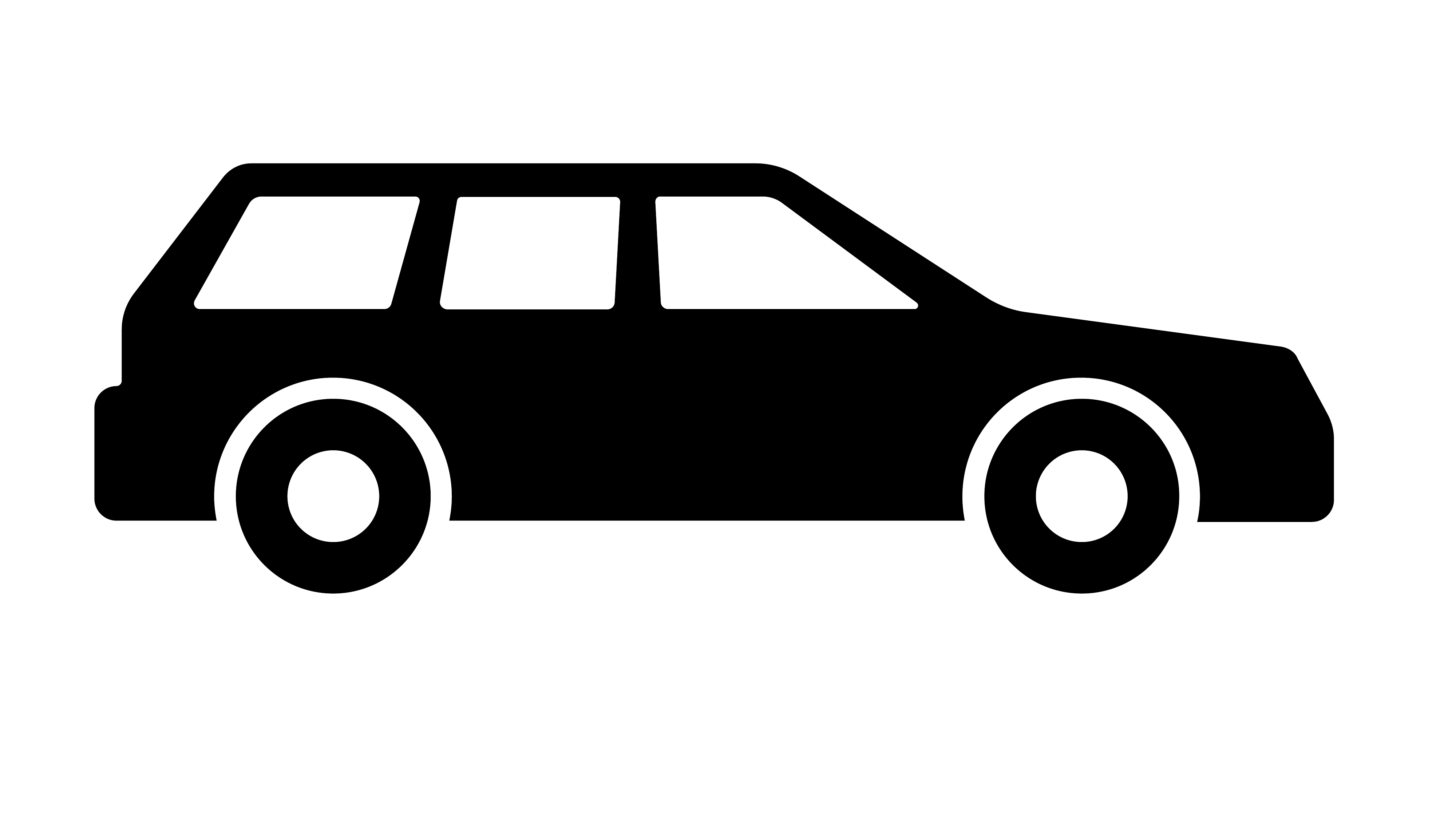}
};
\node[inner sep=0pt, below=0mm of car1] (car1_label)  {
    \footnotesize\sf Agent 1
};
\node[inner sep=0pt, below=0mm of car2] (car2_label)  {
    \footnotesize\sf Agent 2
};
\node[inner sep=0pt, below=0mm of car3] (car3_label)  {
    \footnotesize\sf Agent N
};

\node[draw, inner sep=0pt] at(1.7,0.5) (submap_collection1)  {
    \includegraphics[width=0.1\columnwidth]{figures/img/kitti09/submap1.png}
};
\draw[linestyle_tight] (map1.east) -| ++(1cm,0) |- (submap_collection1.west);
\node[draw, inner sep=0pt, below left=1mm and 1mm of submap_collection1.north east] (submap_collection2)  {
    \includegraphics[width=0.1\columnwidth]{figures/img/kitti09/submap2.png}
};
\draw[linestyle_tight] (map2.east) -| ++(0.9cm,0) |- (submap_collection2.west);
\node[draw, inner sep=0pt, below left=1mm and 1mm of submap_collection2.north east] (submap_collection3)  {
    \includegraphics[width=0.1\columnwidth]{figures/img/kitti09/submap3.png}
};
\draw[arrowstyle_tight] (map3.east) -| ++(0.8cm,0) |- (submap_collection3.west);

\node[inner sep=0pt, above=1mm of submap_collection1] (submap_collection_label)  {
    \footnotesize\sf Meshes
};

\node[processstyle, below=0.3cm of submap_collection2] (tile_assignment)  {
    \footnotesize\sf Tile Assignment
};

\node[processstyle, below=0.3cm of tile_assignment] (point_sampling)  {
    \footnotesize\sf Point Sampling
};

\draw[arrowstyle] (submap_collection2) -- (tile_assignment);
\draw[arrowstyle] (tile_assignment) -- (point_sampling);

\node[posestyle, below=2cm of submap_collection1] (pose_j)  {
    ~~~~$\pose_j$~~~~
};
\draw[linestyle_tight] (map1.east) -| ++(1cm,0) |- (pose_j.west);
\node[posestyle, below left=1mm and 1mm of pose_j.north east] (pose_k)  {
    ~~~~$\pose_k$~~~~
};
\draw[linestyle_tight] (map2.east) -| ++(0.9cm,0) |- (pose_k.west);
\node[posestyle, below left=1mm and 1mm of pose_k.north east] (pose_i)  {
    ~~~~$\pose_i$~~~~
};
\node[inner sep=0pt, below=1mm of pose_i] (pose_label)  {
    \footnotesize\sf Submap Poses
};
\draw[arrowstyle_tight] (map3.east) -| ++(0.8cm,0) |- (pose_i.west);

\draw[arrowstyle] (pose_k) -- (point_sampling);

\node[draw=ccyan, dashed, above right=25mm and 5mm of point_sampling.east, anchor=north west, inner sep=2mm, minimum width=8.25cm, minimum height=3.3cm] (nef)  {};
\node[ccyan, inner sep=0pt, below left=1mm and 1mm of nef.north east] (nef_label)  {
    \footnotesize $\nef(x)$
};
\node[draw=cgreen, dashed, above right=6mm and 1.4cm of point_sampling.east, inner sep=2mm] (featuregrid)  {
\resizebox{1.2cm}{!}{\input{figures/cuboid.tex}
\begin{tikzpicture}[solid]%
    \tikzcuboid{%
        dimx=3,%
        dimy=3,%
        dimz=3,anglez=215,%
        front/.style={draw=cgrey,fill=white},%
        top/.style={draw=cgrey,fill=white},%
        right/.style={draw=cgrey,fill=white},%
    };%
\end{tikzpicture}%
}%
};
\node[draw=cgreen, fill=white, dashed, below left=1mm and 1mm of featuregrid.north east, inner sep=2mm] (featuregrid_2)  {
\resizebox{1.2cm}{!}{\input{figures/cuboid.tex}
\begin{tikzpicture}[solid]%
    \tikzcuboid{%
        dimx=3,%
        dimy=3,%
        dimz=3,anglez=215,%
        front/.style={draw=cgrey,fill=white},%
        top/.style={draw=cgrey,fill=white},%
        right/.style={draw=cgrey,fill=white},%
    };%
\end{tikzpicture}%
}%
};
\node[draw=cgreen, fill=white, dashed, below left=1mm and 1mm of featuregrid_2.north east, inner sep=2mm] (featuregrid_3)  {
\resizebox{1.2cm}{!}{\input{figures/cuboid.tex}
\begin{tikzpicture}[solid]%
    \tikzcuboid{%
        dimx=3,%
        dimy=3,%
        dimz=3,anglez=215,%
        front/.style={draw=cgrey,fill=white},%
        top/.style={draw=cgrey,fill=white},%
        right/.style={draw=cgrey,fill=white},%
    };%
\end{tikzpicture}%
}%
};
\node[cgreen, inner sep=0pt, below right=1mm and 1mm of featuregrid_3.north west] (gridparams_label)  {
    \footnotesize $\Phi_t$
};
\node[inner sep=0pt, above=2mm of featuregrid_2] (featuregrid_label)  {
    \footnotesize\sf Tile Feature-Grids
};

\node[headstyle, below right=2mm and 1.5cm of point_sampling.east] (posenc)  {
    \footnotesize\sf PosEnc
};

\draw[arrowstyle, shorten >=6pt] (point_sampling.east) -| ++(0.75,0) |- node[above,xshift=1mm]{\footnotesize$x$} (featuregrid.west);
\draw[arrowstyle] (point_sampling.east) -| ++(0.75,0) |- node[above,xshift=3mm]{\footnotesize$x$} (posenc.west);

\node[nnstatestyle, above right=3mm and 4cm of point_sampling.east] (feat_aug)  {
    \footnotesize~$\left\{ \features, \embedding \right\}$
};

\draw[revarrowstyle] (feat_aug.west) -| ++(-0.5,0) |- node[right]{\footnotesize$\embedding(x)$} (posenc.east);
\draw[revarrowstyle] (feat_aug.west) -| ++(-0.5,0) |- node[right]{\footnotesize$\features(x,\Phi)$} (featuregrid.east);

\node[headstyle, above right=1mm and 0.7cm of feat_aug.east, minimum width=1cm] (sdf_head)  {
    \footnotesize\sf $\head_\text{Geo}$
};
\node[headstyle, below right=1mm and 0.7cm of feat_aug.east, minimum width=1cm] (sem_head)  {
    \footnotesize\sf $\head_\text{Sem}$
};

\node[fieldstyle, right=0.5cm of sem_head.east, minimum width=1cm] (nef_sem)  {
    \tiny\sf $\nef_\text{sem}(x)$
};

\node[fieldstyle, above right=1mm and 0.5cm of sdf_head.east, minimum width=1cm] (nef_sdf)  {
    \tiny\sf $\nef_\text{sdf}(x)$
};
\node[fieldstyle, below right=1mm and 0.5cm of sdf_head.east, minimum width=1cm] (nef_conf)  {
    \tiny\sf $\nef_\text{conf}(x)$
};
\draw[arrowstyle] (feat_aug.east) -| ++(0.375,0) |- (sdf_head.west);
\draw[arrowstyle] (feat_aug.east) -| ++(0.375,0) |- (sem_head.west);

\draw[arrowstyle] (sem_head.east) -- (nef_sem.west);
\draw[arrowstyle] (sdf_head.east) -| ++(0.2,0) |- (nef_sdf.west);
\draw[arrowstyle] (sdf_head.east) -| ++(0.2,0) |- (nef_conf.west);

\node[processstyle, below=1.1cm of sem_head] (loss)  {
    \footnotesize\sf Loss
};
\draw[arrowstyle] (point_sampling.east) -| ++(0.25,0) |- node[above,xshift=3cm] { {\footnotesize\sf targets }} (loss.west);

\node[processstyle, right=1cm of nef_conf] (mc)  {
    \footnotesize\sf Marching\\
    \footnotesize\sf Cubes
};
\draw[arrowstyle] (nef_sdf.east) -| ++(0.5,0) |- (mc.west);
\draw[arrowstyle] (nef_conf.east) -| ++(0.5,0) |- (mc.west);
\draw[arrowstyle] (nef_sem.east) -| ++(0.5,0) |- (mc.west);

\draw[arrowstyle] (nef_conf.east) -| ++(0.65,0) |- (loss.east);

\node[inner sep=0pt, right=5mm of mc, yshift=-3.5mm] (mesh_fused)  {
    \includegraphics[width=3.5cm]{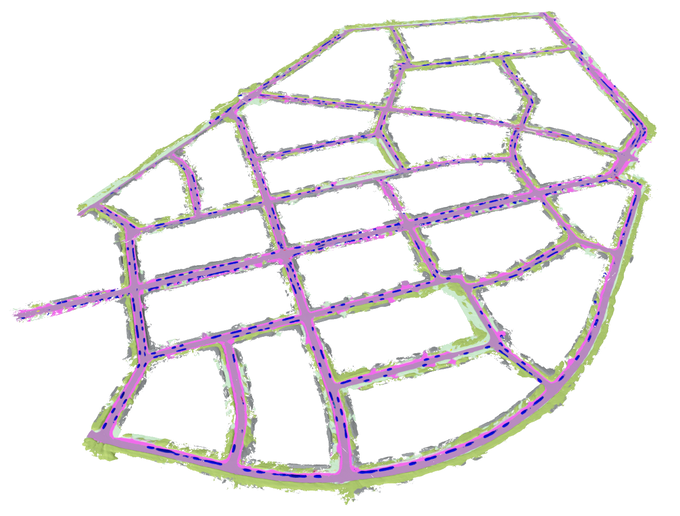}
};
\node[inner sep=0pt, above=1mm of mesh_fused] (mesh_fused_label)  {
    \footnotesize\sf Fused Map Reconstruction
};
\node[draw, inner sep=0pt, below right=-7mm and 1.5cm of mesh_fused.south west, anchor=north east] (mesh_fused_detail)  {
    \includegraphics[width=2cm]{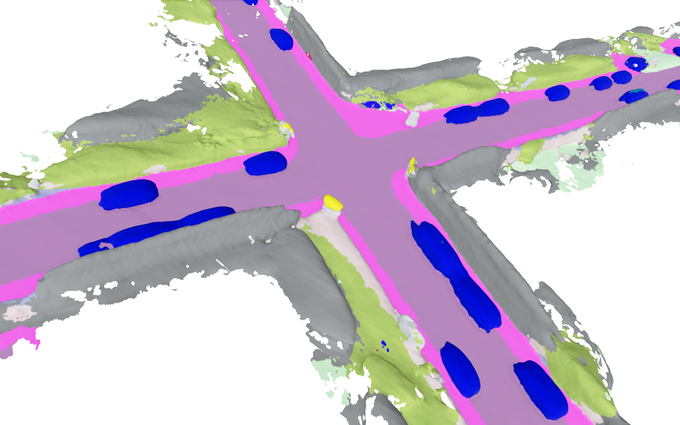}
};

\node[draw, minimum width=5mm, minimum height=3mm, below right=8.75mm and 11.5mm of mesh_fused.north west] (mesh_detail_box) {};
\draw[-] (mesh_detail_box.north west) -- (mesh_fused_detail.north west);
\draw[-] (mesh_detail_box.south east) -- (mesh_fused_detail.south east);

\draw[arrowstyle] (mc.east) -- ([yshift=3.5mm]mesh_fused.west);

\end{tikzpicture}%
    }%
    \caption{System Pipeline overview of our method: We collect meshes and poses $\pose$ for each local agent mapping session, cluster them into different geographic tiles $t$, and use them to supervise the neural semantic signed distance field. This consists of the feature grids $\Phi_t$ for each tile and shared geometry head $\head_\text{Geo}$ predicting SDF $s(x) = \nef_\text{sdf}(x)$ and confidence $c(x) = \nef_\text{conf}(x)$ and semantic head $\head_\text{Sem}$ predicting logits $l(x) = \nef_\text{sem}(x)$, conditioned on grid features $\features$ and point embeddings $\embedding$. We jointly optimize submap poses $\pose_i$, neural field grids $\Phi_t$ and decoder heads $\head$. The final fused map reconstruction can be conveniently extracted from the neural field using Marching Cubes.}%
    \label{fig:cloudmap:pipeline}%
\end{figure*}%

\section{Related Work}
Mapping has been a central topic in the robotics community for decades, with an abundance of work in the realm of Simultaneous Localization and Mapping (SLAM). We focus our review on the most related works in the fields of maps for autonomous vehicles, dense reconstruction and multi-robot mapping using vision sensors.

\subsection{Mapping for Autonomous Vehicles}
Crowd sourced mapping has seen a surge in interest in recent years, particularly for the application of autonomous vehicles.
However, most works in the field focus on specific map features related to road networks, such as lane markings or traffic signs \cite{Dabeer2017}, which are light-weight and easy to gather in a crowd-sourced fashion.
A special focus in these works has been put on detailed mapping of the geometry \cite{Herb2019iros,Qin2021}, topology \cite{Pannen2019b} or dealing with change detection and update \cite{Pannen2019a,Kim2021a} for lane markings.
However, these works focus only on specific road environments and offer a limited spatial understanding due to the extreme level of environment abstraction, which restricts its usefulness for tasks like localization or augmented reality.
In contrast, Cheng \etAl \cite{Cheng2022} use semantically labeled sparse points from visual odometry that are more generic, but still sparse.

\subsection{Dense Map Reconstruction}

Structure-from-Motion (SfM) methods such as COLMAP \cite{Schonberger2016} are well established for large-scale sparse and dense reconstruction from image-collections. While SfM methods produce high-quality dense reconstructions, they demand significant computational resources for multi-view stereo and need to collect large amounts of image data.
As an alternative, fast online reconstructions using surfels \cite{Wang2019a} or meshes \cite{Herb2021a} have been proposed for robotic applications.

While the aforementioned approaches represent 3D structure \emph{explicitly} as points, meshes or surfels, implicit methods leveraging Signed Distance Fields (SDF) as representation have been used for high-quality 3D reconstructions.
Hashed voxel-grid structures are commonly used to reconstruct an SDF \cite{oleynikova2017voxblox}, especially for small-scale environments, with extensions towards larger-scale \cite{Reijgwart2020} and semantic information \cite{Rosinol2019a} enabling map reconstructions for street-scale maps \cite{Hu2023}.
However, very detailed large-scale outdoor reconstructions using voxel-representations are limited by prohibitively high memory requirements arising from small voxel-sizes.

Recently, Neural Implicit Representations pioneered by NeRFs \cite{Mildenhall2020a} for view synthesis and DeepSDF \cite{Park2019} for geometry reconstruction have shown groundbreaking results by learning properties of a specific scene such as density, color or signed distance as a Multi-Layer Perceptron (MLP) that can be queried for every 3D point in the scene.
Early works on large-scale scene reconstruction using images \cite{Tancik2022} and LiDAR \cite{Rematas2022} show impressive results, but are slow to train.
Using hashed feature-grids \cite{Mueller2022instantngp} or OcTrees \cite{Takikawa2021,Zhong2022shinemapping} for storing scene features in combination with a shallow MLP head has enabled dramatic speedups for neural field training and inference, especially on large-scale scenes.
In addition to just reconstructing the scene content, joint optimization with input poses for cameras \cite{Lin2021} and LiDARs \cite{Deng2023} has been proposed to overcome the need for posed sensor data.

\subsection{Multi-Robot Mapping}
While a majority of SLAM works address the problem of single-robot mapping, many practical applications require the use of multiple agents working together in a centralized or decentralized setup to map larger environments or keep them up to date.

ORBSLAM-Atlas \cite{Elvira2019} and CCM-SLAM \cite{Schmuck2019} both build on top of ORB-SLAM \cite{Mur-Artal2015a} to extend it to a centralized multi-robot mapping system, with a focus on map accuracy and inter-agent communication respectively.
\texttt{maplab 2.0} \cite{Cramariuc2023} is another framework for multi-robot multi-session mapping that follows a centralized approach, which supports various modalities including visual keypoints, inertial and GNSS data.
However, due to its sparse map representation, all of these approaches are primarily suited to build maps for (re-)localization tasks.

Multiple works have expanded the multi-robot SLAM problem to incorporate semantic and dense map information in various ways.
Kimera-Multi \cite{Chang2020} extends Kimera \cite{Rosinol2019a} into a decentralized mapping system, however the multi-robot part focuses on a joint sparse pose-graph optimization, while each agent keeps and refines its dense mesh reconstruction, such that no dense map information is created collaboratively.

Golodetz \etAl \cite{Golodetz2018} use a centralized approach, where agents transmit RGB-D data to a powerful server that performs map reconstruction for all robots.
In \cite{Dubois2020} a decentralized multi-robot mapping system is presented that uses stereo cameras to reconstruct TSDF-based submaps that are shared among robots and geometrically aligned as part of a pose-graph optimization. Both such approaches are hard to scale to large fleets of robots due to large amount of resources requirement for data exchange and storage.
Coxgraph \cite{Liu2021} uses a similar approach, but compresses the TSDF maps into meshes for transmission to a central mapping server, where the SDF is recovered from the mesh.
We also adopt meshes as exchange format for map data from local agents to a centralized server in our system, enabling lightweight data transmission and compatibility with a wide variety of local map reconstruction methods.
In contrast to existing approaches, we employ a neural implicit representation to fuse the map data contributed from each agent.
This allows us to optimize geometry, semantics and poses jointly without the need for specific localization landmarks and facilitate large-scale high-fidelity map reconstruction.

\section{Approach}
\subsection{System Overview}
In \refFigure{fig:cloudmap:pipeline} we show an overview of our pipeline, which is made up of the local map generation in each agent, which we describe in \refSection{sec:cloudmap:submap_gen}, and the main focus of our work, the fusion component, explained in \refSection{sec:cloudmap:nef}.
In our system, each mapping agent produces local submaps from onboard sensory data, which are uploaded to a central mapping server by each agent as a semantically labelled mesh. Afterwards, we use all submaps to supervise the neural field to produce a fused semantic signed distance field. 

\subsection{Submap Generation}
\label{sec:cloudmap:submap_gen}
On each agent, we run a local dense semantic SLAM to generate mapping data for the environment that the agent explores.
In this work we focus on crowd-sourced mapping using low-cost monocular-only cameras, but our method is applicable to any dense semantic SLAM system, e.g. also LiDAR based methods.
We use ORB-SLAM2 \cite{Mur-Artal2017a} in monocular configuration to generate local vehicle odometry for each agent.
To demonstrate the applicability to different local mapping systems, we show results using a lightweight mesh mapping from semantic image segmentation and monocular depth \cite{Herb2021a} as well as single RGB semantic scene completion \cite{Cao2022} for dense mesh reconstruction.
In sum, each agent transmits local vehicle odometry as well as corresponding GPS measurements and a number of submaps encoded as triangle mesh, with vertices, triangles, and semantic labels for each face or vertex.

\subsection{Neural Map Fusion}
\label{sec:cloudmap:nef}
After collecting all submaps from each agent-session, we first cluster all submaps into fixed-size square tiles using the provided coarse GPS measurements. This ensures scalability to larger scenes as only a subset of tiles-of-interest can be loaded for training or inference and is comparable to voxel-hashing techniques in traditional voxel-based surface reconstruction.

\subsubsection{Neural Semantic Signed Distance Field}
We represent the fused map as a neural field $\nef(x)$, a neural network which learns to predict the geometry and semantics for arbitrary input sampling points $x \in \real^3$.
As in other neural surface reconstruction methods \cite{Zhong2022shinemapping,Deng2023,Li2023}, we use a signed distance field denoted as $s(x) = \nef_\text{sdf}(x) : \real^3 \mapsto \real$ to model the geometry, where the zero-levelset defines the surface implicitly, from which a dense surface mesh can be extracted by applying marching cubes on a regularly sampled grid.
We use dedicated feature-grids $\Phi_t$ for each spatial tile $t$ and combine these with shallow multi-layer perceptrons (MLPs) as output heads $\head$ shared across all tiles, which ensures scalability while keeping training time low.
In a first step, we interpolate point-wise features $\features(x, \Phi_t)$  tri-linearly from the corresponding grid, which are stacked with the positional encodings $\embedding(x)$ of the input coordinates $x$ and feed them into the prediction heads $\head_\text{head}\left(\left\{ \features(x, \Phi_t), \embedding(x) \right\}\right) = \left\{\nef_\text{pred}(x), \ldots \right\}$.
While positional encoding is not commonly used in existing neural fields conditioned on 3D grid features\cite{Mueller2022instantngp,Zhong2022shinemapping}, we found that it improves the pose optimization (see \refSection{sec:cloudmap:ablation:alignment}).

\paragraph{Feature Grids}
We adopt OcTrees \cite{Takikawa2021} and HashGrids \cite{Mueller2022instantngp} as sparse and memory-efficient feature-grid structures, which are commonly used in neural fields to enable large-scale reconstruction.
For both grids, features are stored in a lookup-table (codebook) with the mapping of point coordinates to features being defined by either the voxel-index inside the OcTree or by a spatial hash function.
We evaluate the suitability of both structures for our fusion approach in \refSection{sec:cloudmap:eval}.

\paragraph{Geometry Head}
Our geometry MLP head $\head_\text{Geo}(x) = \left\{s(x), c(x)\right\}$ primarily predicts the signed distance field $s(x) = \nef_\text{sdf}(x)$.
In addition, it outputs a confidence score $c(x) = \nef_\text{conf}(x) : \real^3 \mapsto [0, 1]$, which represents the likelihood of the surface or objects existence.
While the signed distance in an SDF represents the estimated distance towards the closest object surface, there is no notion of existence measure associated to it.
When using data from multiple, contradictory sessions with noisy or spurious measurements, it is crucial to denote how reliable a surface prediction is.
For instance, some temporary object might be detected only by one or few agents and this should be modelled accordingly by the neural field.

\paragraph{Semantic Head}
In addition to the geometry head, we use a dedicated semantic head $\head_\text{Geo}(x) = \left\{ l(x) \right\}$, which predicts semantic logits $l(x) = \nef_\text{sem}(x) : \real^3 \mapsto \real^K$.
From this, we derive class scores $L(x_\text{surf}) = \sigma\left(l(x_\text{surf})\right) \in [0,1]^K$ with $\sigma$ being the \emph{softmax} operation.
We found that using a dedicated decoder MLP for semantics slightly improved the reconstruction quality.

\subsubsection{Pose Optimization}

We expect the input submaps contributed by each agent to be geo-localized, as GPS sensors are commonly available.
However, the typical error of GPS measurements is relatively large compared to the required accuracy needed for centimeter-accurate mapping.
In order to overcome this, we optimize both the neural field parameters \emph{and} the input poses jointly to produce optimal reconstruction results.
For each input submap $i$, we estimate a pose $\pose_i$ in global coordinates, such that local submap coordinates $x_i$ can be transformed to global coordinates $\tilde{x}_i = \transform(\pose_i, x_i)$.
Each pose $\pose_i \in \se(3)$ is represented as Lie algebra to facilitate optimization and is initialized with the GPS measurement $\hat{\pose}_{\text{GPS}, i}$.

\subsection{Map Reconstruction \& Alignment}

\subsubsection{Supervision Point Sampling}

To fuse the submaps into a common environment reconstruction, we train the neural field using individual points, sampled from the input meshes.
For each batch, we sample points $x_i$ uniformly distributed on the surface of the input meshes $i$ and transform them to the global frame using corresponding poses $\tilde{x}_i=\transform(\pose_i, x_i)$ to obtain surface samples $\hat{x}_\text{surf} = \left\{\tilde{x}_0, \tilde{x}_1, \ldots\right\}$.
From these we compute target samples $x_\text{surf}=\hat{x}_\text{surf}+\hat{s}\cdot\hat{n}_\text{surf}$ around the surface at an offset $\hat{s} \sim \normal(0, \sigma_s)$ with $\sigma_s = 0.05m$ along the surface normal $\hat{n}_\text{surf}$ such that $\hat{s}$ represents the target signed distance for $x_\text{surf}$.
We also sample the same number of points $x_\text{space}$ uniform randomly in all input submap bounding boxes for regularization and confidence priors.

\subsubsection{Losses}

Our primary objective is the surface reconstruction, for which we regress SDF value of the target points $x_\text{surf}$ sampled around the surface
\begin{equation}
    \loss_\text{sdf}(s(x_\text{surf})) = \left(s(x_\text{surf})-\hat{s}\right)^2
\end{equation}
with $\hat{s}$ being the offset distance of the sampled point along the surface normal.
We further regress the predicted surface normals $n(x_\text{surf})$ against the normals sampled from the input meshes $\hat{n}_\text{surf}$ for each point by computing the normal prediction as $n(x) = \frac{\partial}{\partial x} s(x)$ using the numerical gradient method from \cite{Li2023} and formulate the loss as
\begin{equation}
    \loss_\text{norm}(n(x_\text{surf})) = \left\lVert n(x_\text{surf}) - \hat{n}_\text{surf} \right\rVert^2.
\end{equation}
For semantics, we use a standard cross-entropy (CE) loss on predicted class scores $L(x_\text{surf}) \in [0,1]^K$ and target labels $\hat{L}_\text{surf} \in \nat$:
\begin{equation}
    \loss_\text{sem}(L(x_\text{surf})) = \mathrm{CE}(L(x_\text{surf}), \hat{L}_\text{surf}).
\end{equation}
The target labels $\hat{L}_\text{surf}$ for each point $x_\text{surf}$ sampled around the surface correspond to the label of the input mesh face the surface point was sampled from.
To regularize the SDF, we add an eikonal loss \cite{Gropp2020Eikonal} to all sampled surface and volume points:
\begin{equation}
    \loss_\text{eik}(x_\text{surf,space}) = \left(\left\lVert \frac{\partial}{\partial x_\text{surf,space}} s(x_\text{surf,space}) \right\rVert_2-1\right)^2
\end{equation}
enforcing the gradient-vector of the SDF w.r.t. input coordinates to be close to unit-norm.
We supervise the confidence output using a binary cross-entropy (BCE) loss to predict $1$ for valid surfaces $x_\text{surf}$ and $0$ for non-surface areas $x_\text{space}$: 
\begin{align}
    \loss_\text{conf}(c(x_\text{surf}), c(x_\text{space})) &= \mathrm{BCE}(c(x_\text{surf}), 1) \nonumber \\
    &+ \mathrm{BCE}(c(x_\text{space}), 0)
\end{align}
We add an odometry loss to stabilize the pose optimization by enforcing that the relative pose $\pose_{i\to{i+1}} = \pose^{-1}_{i+1}\pose_{i}$ of consecutive submaps $i, i+1$ is consistent with their measured odometry $\hat{\pose}_{\text{odom},i\to{i+1}}$
\begin{equation}
    \loss_\text{odom}(\pose_{i\to i+1}) =
    w_\text{odom} \hat{\pose}^{-1}_{\text{odom},i\to i+1}\pose_{i\to i+1}
\end{equation}
where $w_\text{odom}$ is weight-vector to model the uncertainty in the odometry.
Our total loss is therefore composed as
\begin{align}
    \loss_\text{total} &= \loss_\text{sdf} + \lambda_\text{norm}\loss_\text{norm} + \lambda_\text{sem}\loss_\text{sem} \nonumber \\
    &+\lambda_\text{eik}\loss_\text{eik} + \lambda_\text{conf}\loss_\text{conf} + \lambda_\text{odom}\loss_\text{odom}
\end{align}.
\section{Evaluation}
\label{sec:cloudmap:eval}
In the following, we present an evaluation of our approach on two different datasets using different methods for submap-mesh generation.

\subsection{Experimental Setup}
We implemented the proposed approach in PyTorch using the feature HashGrid and OcTree implementation of Kaolin Wisp \cite{KaolinWispLibrary}, with modifications to ensure gradient propagation for pose optimization in PyTorch.
We train all tiles of size $128m\times128m$ jointly, randomly selecting one tile per iteration, using AdamW with learning rate $10^{-2}$, weight decay of $10^{-2}$ and LR decay of $10^{-0.001\frac{\mathrm{iter}}{\mathrm{\#tiles}}}$ for grid and model parameters.
Pose parameters use learning rates of $10^{-2}$ and $10^{-4}$ for translational and rotational components respectively, without weight decay or LR decay.
We train for an average of $500$ iterations per tile with $125000$ surface and volume samples each per batch.
For the HashGrid, we use $16$ levels of feature size $2$, with grid sizes from $2^4$ to $2^{11}$ and a codebook size of $2^{N_C} = 2^{16}$.
Our OcTree configuration uses $4$ feature-levels with leaf-voxel size of $0.5m$ and feature size $N_F=4$ per level.
We further apply progressive level-of-detail (LOD) optimization \cite{Li2023}, where we activate only the coarser half of feature-levels in the beginning and incrementally add one additional layer every $L$ iterations.
Inactive levels are masked out to zero.
We use separate MLP heads for geometry and semantics with $2$ hidden layers and $128$ neurons each.
All experiments were conducted on an NVIDIA V100 with 32GB VRAM, with training taking approximately 90 seconds per tile.
We use $\lambda_\text{norm} = \lambda_\text{sem} = \lambda_\text{conf} = \lambda_\text{odom} = 1, \lambda_\text{eik} = 0.1$ and a confidence threshold $c_\text{th}$ of $0.7$ when extracting meshes for all datasets.
\begin{table*}[!th]%
    \setlength{\tabcolsep}{5.3pt}%
    \centering%
    \caption{Alignment and reconstruction results overview for different combinations of pose-alignment and map reconstructions. Scores represent translational [m] and rotational [deg] RMSE as well as geometric (geo) and mean semantic (sem) F-Score for all datasets. Highlighted results denote \textbf{best} and \textit{second-best} per column.}%
    \label{tab:cloudmap:eval:overview}%
    \begin{tabular}{p{2cm}lrrrrrrrrrrrr}%
    \toprule%
    \multirow{2}{*}{Method} & & \multicolumn{4}{c}{HD-Map} & \multicolumn{4}{c}{Kitti360 06} & \multicolumn{4}{c}{Kitti360 09} \\%
    \cmidrule(lr){3-6}\cmidrule(lr){7-10}\cmidrule(lr){11-14}%
                          &         &   trans $\downarrow$ &   rot $\downarrow$ &   geo $\uparrow$ &   sem $\uparrow$ &  trans $\downarrow$ &    rot $\downarrow$ &   geo $\uparrow$ &   sem $\uparrow$ &  trans $\downarrow$ &   rot $\downarrow$ &  geo $\uparrow$ &  sem $\uparrow$ \\%
    \midrule%
    \multirow{4}{*}{Poisson}      & ICP &                      1.029 &                       1.104 &            0.333 &            0.152 &                      3.774 &                       3.466 &            0.108 &            0.051 &                      4.044 &                       3.485 &            0.089 &            0.047 \\%
                                  & ICP+Sem &                      1.016 &                       1.101 &            0.361 &            0.169 &                      3.818 &                       3.513 &            0.107 &            0.049 &                      3.997 &                       3.524 &            0.086 &            0.053 \\%
                                  & GICP &                      1.256 &                       1.432 &            0.314 &            0.148 &                      2.197 &                       3.685 &            0.100 &            0.059 &                      2.039 &                       3.608 &            0.127 &            0.084 \\%
                                  & GICP+Sem &                      1.227 &                       1.425 &            0.277 &            0.138 &                      2.163 &                       3.631 &            0.103 &            0.060 &                      2.086 &                       3.618 &            0.125 &            0.091 \\%
    \midrule%
    \multirow{2}{*}{Neural Field (ours)} & OcTree                    &           $\mathit{0.830}$ &            $\mathit{1.030}$ & $\mathit{0.552}$ & $\mathit{0.273}$ &           $\mathit{1.710}$ &            $\mathbf{2.040}$ & $\mathbf{0.168}$ & $\mathbf{0.104}$ &           $\mathbf{1.057}$ &            $\mathbf{1.568}$ & $\mathbf{0.431}$ & $\mathbf{0.280}$ \\%
                                         & HashGrid                  &           $\mathbf{0.810}$ &            $\mathbf{0.924}$ & $\mathbf{0.605}$ & $\mathbf{0.294}$ &           $\mathbf{1.603}$ &            $\mathit{2.069}$ & $\mathit{0.159}$ & $\mathit{0.099}$ &           $\mathit{1.076}$ &            $\mathit{1.597}$ & $\mathit{0.410}$ & $\mathit{0.266}$ \\%
\bottomrule%
\end{tabular}%
\end{table*}%
%
\subsection{Datasets}%
\paragraph{KITTI360}
KITTI360 \cite{Liao2022} is an outdoor driving dataset covering suburban areas with front-facing stereo camera and semantically labeled LiDAR point clouds as ground truth.
While the dataset only contains single traversals of the same scene, the trajectories are very loopy, often covering the same area in forward and backward direction.
To further increase the coverage, we use only monocular camera data and treat left and right stereo camera as independent traversals.
To generate semantic input meshes to be fused, we use MonoScene \cite{Cao2022} to predict a semantic occupancy map, which we convert into a triangle mesh.
We choose sequences 6 and 9 for evaluation and use the rest to fine-train MonoScene on KITTI360, initialized from the official model trained on SemanticKITTI.
We use the SLAM-optimized frame poses as ground-truth and the raw OXTS measurements as GPS poses, which we disturb using artificial gaussian noise since OXTS already provides highly accurate GPS poses.
We use every 5th keyframe for mapping and group consecutive frames into chunks of length 5.

\paragraph{HD Map Dataset}
In addition, we use a private dataset that features 7 traversals of an urban route of around 3 km in length in forward and backward direction using a single front-facing camera mounted in the vehicle and standard GPS.
As ground truth, we use a manually annotated high-definition map and global pose measurements from RTK-GPS. We use the method in \cite{Herb2021a} to generate lightweight mesh submaps of 10 keyframes each.

\subsection{Baseline}
We compare our system against traditional map-alignment methods ICP \cite{Besl1992} and Generalized ICP \cite{Segal2010}, which we apply in a factor-graph with additional pose constraints as in our approach. For each submap, we select 3 neighboring submaps with the highest spatial overlap.
On each submap-mesh we sample $500$ points uniformly and compute the (G)ICP-loss over all submap-pairings. We additionally implement a semantic-mode that only associates points with matching semantic label.
We optimize the alignment loss with robust Huber kernel using the same optimizer as our neural field, and use the aligned poses as input for the map reconstruction.

For map reconstruction, we build a simple but strong baseline by applying state-of-the-art mesh reconstruction from oriented points using Poisson Surface Reconstruction \cite{Kazhdan2013} in COLMAP \cite{Schonberger2016} on 1000000 points with surface normals sampled from all input submaps for each tile.
We found that using more sample points did not improve the reconstruction while increasing processing time significantly.
After mesh-fusion, we transfer labels from the input meshes by selecting the most-common label of the 5 closest points to each mesh face.

\subsection{Evaluation Metrics}
\subsubsection*{Map Reconstruction}
To evaluate the quality of the reconstructed map, we adopt precision and recall metrics from \cite{Knapitsch2017}, measuring accuracy and completeness of the reconstruction respectively, and the F$_1$-Score as harmonic mean of both as geometric (\emph{geo}) reconstruction score.
We additionally compute the corresponding semantic (\emph{sem}) scores as average over per-class F-scores.
We choose a distance threshold of $d=20cm$.

\subsubsection*{Pose Alignment}
For pose optimization, we report the absolute root-mean squared (RMSE) translation and rotation errors between estimated poses and corresponding ground-truth poses after rigid alignment of the entire trajectory to the ground-truth.%
\begingroup
\captionsetup[subfloat]{farskip=5pt}
\begin{figure*}[!th]%
    \centering%
    \subfloat{\includegraphics[width=0.19\textwidth]{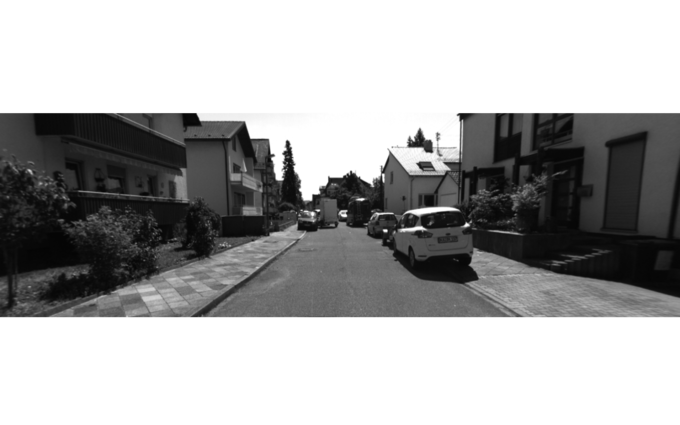}} ~%
    \subfloat{\includegraphics[width=0.19\textwidth]{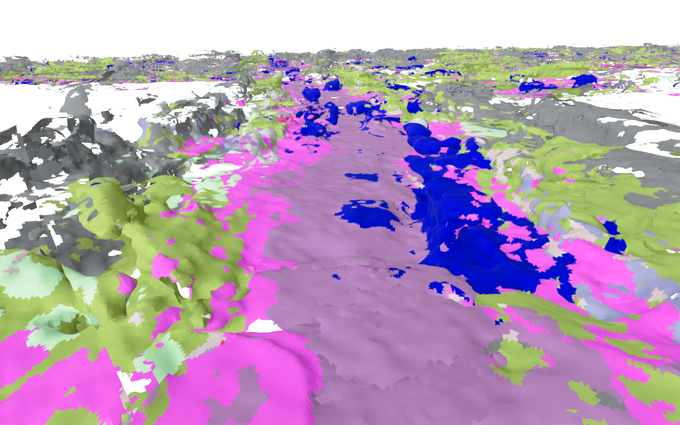}} ~%
    \subfloat{\includegraphics[width=0.19\textwidth]{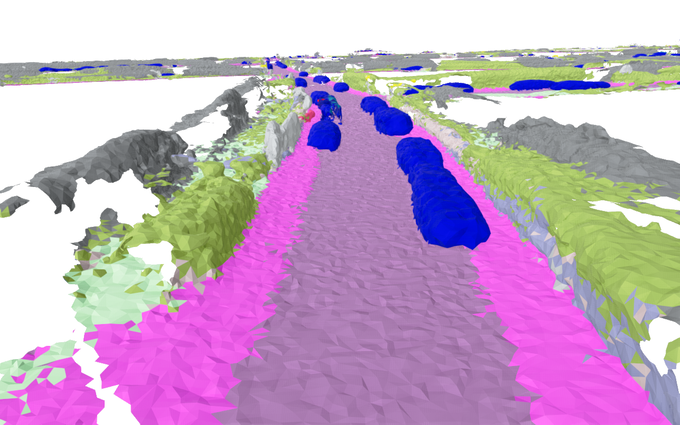}} ~%
    \subfloat{\includegraphics[width=0.19\textwidth]{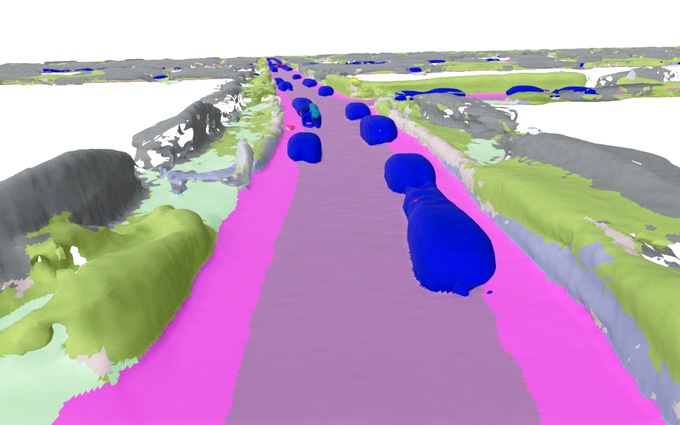}} ~%
    \subfloat{\includegraphics[width=0.19\textwidth]{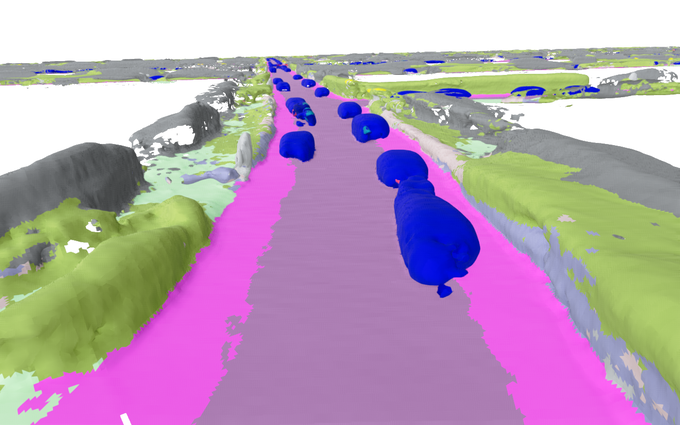}}\\%
    \setcounter{subfigure}{0}%
    \subfloat[Scene image]{\includegraphics[width=0.19\textwidth]{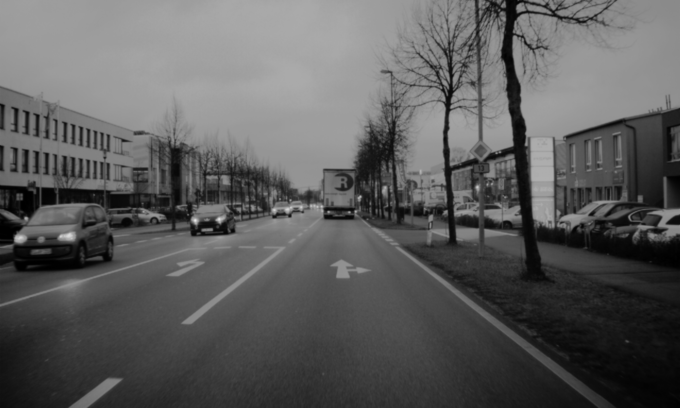}\label{fig:cloudmap:qual:frame}} ~%
    \subfloat[HashGrid GPS (init)]{\includegraphics[width=0.19\textwidth]{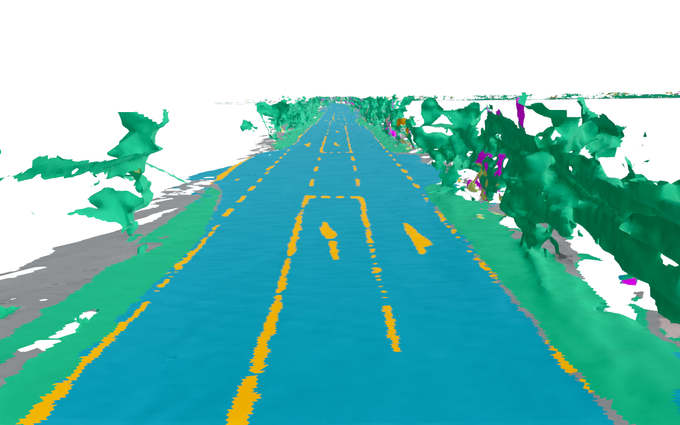}\label{fig:cloudmap:qual:ngp_gps}} ~%
    \subfloat[Poisson+ICP]{\includegraphics[width=0.19\textwidth]{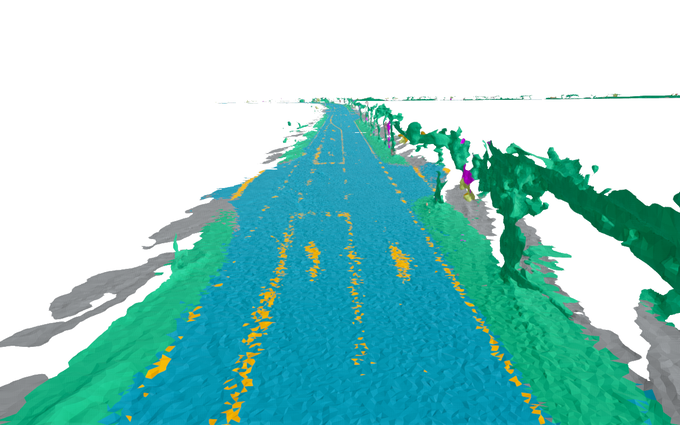}\label{fig:cloudmap:qual:poisson_icp}} ~%
    \subfloat[HashGrid opt (ours)]{\includegraphics[width=0.19\textwidth]{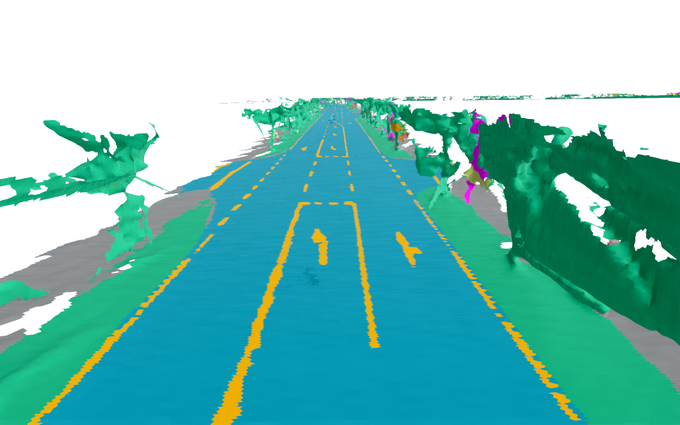}\label{fig:cloudmap:quasl:ngp_num}} ~%
    \subfloat[HashGrid GT (oracle)]{\includegraphics[width=0.19\textwidth]{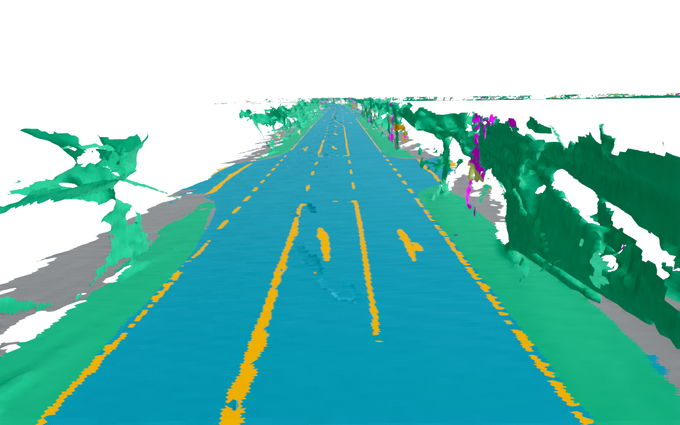}\label{fig:cloudmap:qual:ngp_world}}%
    \caption{Qualitative reconstruction results for KITTI360 (top) and HD-Map (bottom).}%
    \label{fig:cloudmap:results_qualitative}%
\end{figure*}%
\endgroup%
\subsection{Quantitative Results}%

In \refTable{tab:cloudmap:eval:overview} we provide an overview of alignment and reconstructions results for different combinations of pose alignment and geometry reconstruction.
We notice that our proposed method, which optimizes poses and reconstruction jointly, achieves significantly better alignment and reconstruction results over the baseline across all datasets.
For the ICP-based alignment, we found that point-to-point ICP and generalized ICP perform overall similarly with slight benefits for either on the different datasets. Including semantic information has only marginal effect on the alignment.
In our neural field reconstruction, we find that HashGrid and OcTree representations achieve overall similar alignment and reconstruction results.

\subsection{Qualitative Results}
\refFigure{fig:cloudmap:results_qualitative} shows qualitative reconstruction results of our approach.
We compare neural reconstructions using noisy GPS, optimized alignment and oracle (GT) poses as well as Poisson reconstruction.
While direct reconstruction from noisy GPS leads to inferior results, joint neural alignment and reconstruction and factor-graph aligned Poisson meshing achieve visual quality close to reconstruction with oracle poses.
We note that Poisson reconstruction lacks some detail in the reconstruction, e.g. road markings, while neural reconstruction is crisper.
We present additional qualitative results in our supplementary video.

\subsection{Ablation Studies}
In the following we present multiple ablation studies to demonstrate the effect of individual components of our system.
\subsubsection{Submap Alignment}
\label{sec:cloudmap:ablation:alignment}
In \refTable{tab:cloudmap:ablation:alignment} we demonstrate the effect of different alignment configurations.
We observe that incorporating relative pose constraints is important to stabilize convergence.
Furthermore, both the added positional encoding and the incremental level-of-detail reduce the alignment error individually for HashGrids and result in the best overall configuration when used jointly for both feature representations.

\begin{table}[th]
    \centering
    \caption{
        Alignment results on HD-Map dataset for neural field (NeF) optimization with relative pose constraints (rel), positional encoding (pe) and progressive level-of-detail (lod).
        Scores represent mean$\pm$std. of translational [m] and rotational [deg] RMSE over 3 runs.
    }
    \label{tab:cloudmap:ablation:alignment}
    \setlength{\tabcolsep}{3pt}
    \begin{tabular}{lcccrrrr}
        \toprule
        \multicolumn{4}{c}{Pose Opt.} & \multicolumn{2}{c}{HashGrid} & \multicolumn{2}{c}{OcTree} \\
        \cmidrule(lr){1-4}\cmidrule(lr){5-6}\cmidrule(lr){7-8}
        align                  & rel    & pe     & lod    & trans $\downarrow$ & rot  $\downarrow$ & trans  $\downarrow$ & rot  $\downarrow$ \\
        \midrule
        \multicolumn{4}{l}{none}                         &                1.42 &                     2.15 &                     1.42 &                     2.15 \\
        \midrule
        \multirow{5}{*}{NeF}  &        &        &        &            1.21$\pm$0.01 &            2.03$\pm$0.01 &            1.13$\pm$0.01 &            1.93$\pm$0.02\\
                              & \cmark &        &        &            1.06$\pm$0.02 &            1.32$\pm$0.03 &            0.96$\pm$0.00 &            0.99$\pm$0.02\\
                              & \cmark & \cmark &        &            0.97$\pm$0.01 &            1.08$\pm$0.01 &            0.88$\pm$0.00 &            1.04$\pm$0.00\\
                              & \cmark &        & \cmark &            0.85$\pm$0.01 &            1.14$\pm$0.01 &            0.88$\pm$0.01 & $\mathbf{0.94}$$\pm$0.01\\
                              & \cmark & \cmark & \cmark & $\mathbf{0.81}$$\pm$0.00 & $\mathbf{0.94}$$\pm$0.01 & $\mathbf{0.84}$$\pm$0.01 &            1.02$\pm$0.02\\
        \bottomrule
    \end{tabular}
\end{table}

\subsubsection{Reconstruction}
In addition to joint alignment and reconstruction, we also evaluate the reconstruction results with various fixed input poses using GPS (input), aligned using neural optimization and ground-truth poses in \refTable{tab:cloudmap:ablation:recon}.
We observe that all reconstruction methods perform similarly for semantic scores, while Poisson reconstruction achieves slightly higher scores in the geometric reconstruction, as long as input poses are of high quality. 

\begin{table}[th]
    \centering
    \caption{
        Reconstruction scores for different reconstruction methods for fixed submap poses from varying sources on HD-Map dataset. \emph{Aligned} denotes reconstruction from fixed input poses pre-optimized using Neural Field optimization.
    }
    \label{tab:cloudmap:ablation:recon}
    \setlength{\tabcolsep}{6pt}
    \begin{tabular}{lrrrrrr}
        \toprule
        \multirow{2}{*}{Recon} & \multicolumn{2}{c}{GPS} & \multicolumn{2}{c}{Aligned} & \multicolumn{2}{c}{Ground-Truth} \\
        \cmidrule(lr){2-3}\cmidrule(lr){4-5}\cmidrule(lr){6-7}
                              & geo  $\uparrow$ & sem  $\uparrow$ & geo  $\uparrow$ & sem  $\uparrow$ & geo  $\uparrow$ & sem  $\uparrow$ \\
        
        \midrule
        Poisson                   &            0.140 &            0.067 & $\mathbf{0.636}$ & $\mathbf{0.329}$ & $\mathbf{0.737}$ & $\mathbf{0.408}$ \\
        OcTree                    & $\mathbf{0.177}$ & $\mathbf{0.086}$ &            0.610 &            0.309 &            0.712 &            0.376\\
        HashGrid                  &            0.122 &            0.063 &            0.613 &            0.300 &            0.721 &            0.392 \\
        \bottomrule
    \end{tabular}
\end{table}

\begin{figure}[t!]%
    \subfloat[\# Sessions]{%
        \pgfplotstableread{figures/eval/data/sessions_gps/ablation_sessions.dat}\ablationsessiondata%
\begin{tikzpicture}[inner frame sep=0, baseline]%
    \begin{axis}[%
        PlotStyleHalfColumn,%
        trim axis left,%
        width=0.44\columnwidth,%
        height=0.44\columnwidth/\goldenratio,%
        xtick=data,%
        ymax=0.36,%
        ymin=0.08,%
        ytick distance=0.04,%
        y tick label style={%
        /pgf/number format/.cd,%
            fixed,%
            fixed zerofill,%
            precision=2,%
        /tikz/.cd%
        },%
        legend columns=3,%
        legend style={%
            at={(0.5,1.03)},%
            anchor=south,%
            draw=none,
            nodes={scale=0.75, transform shape}%
        }%
    ]%
        \addplot+ table[x=sessions,y=fscore_mean]{\ablationsessiondata};%
        \addplot+ table[x=sessions,y=precision_mean]{\ablationsessiondata};%
        \addplot+ table[x=sessions,y=recall_mean]{\ablationsessiondata};%
        \legend{F-Score, Precision, Recall}%
    \end{axis}%
\end{tikzpicture}%
%
        \label{fig:cloudmap:ablation:sessions:count}%
    }%
    \subfloat[Mapping data \lbrack MB/km\rbrack]{%
        \pgfplotstableread{figures/eval/data/sessions_gps/ablation_sessions-1.dat}\ablationsessiondataone%
\pgfplotstableread{figures/eval/data/sessions_gps/ablation_sessions-5.dat}\ablationsessiondatafive%
\pgfplotstableread{figures/eval/data/sessions_gps/ablation_sessions-10.dat}\ablationsessiondataten%
\pgfplotstableread{figures/eval/data/sessions_gps/ablation_sessions-20.dat}\ablationsessiondatatwenty%
\begin{tikzpicture}[inner frame sep=0, baseline]%
    \begin{axis}[%
        PlotStyleHalfColumn,%
        trim axis right,%
        width=0.44\columnwidth,%
        height=0.44\columnwidth/\goldenratio,%
        xmin=5,%
        xmax=100,%
        xmode=log,%
        log ticks with fixed point,%
        ymax=0.36,%
        ymin=0.08,%
        ytick distance=0.04,%
        yticklabels={},%
        legend columns=4,%
        legend style={%
            at={(0.5,1.03)},%
            anchor=south,%
            draw=none,%
            nodes={scale=0.75, transform shape}%
        }%
    ]%
        \pgfplotsset{cycle list shift=3}%
        \addplot+ table[x=sessions_total_mb_km,y=fscore_mean]{\ablationsessiondataone};%
        \addplot+ table[x=sessions_total_mb_km,y=fscore_mean]{\ablationsessiondatafive};%
        \addplot+ table[x=sessions_total_mb_km,y=fscore_mean]{\ablationsessiondataten};%
        \addplot+ table[x=sessions_total_mb_km,y=fscore_mean]{\ablationsessiondatatwenty};%
        \legend{1,5,10,20 Frames/SM}%
    \end{axis}%
\end{tikzpicture}%
%
        \label{fig:cloudmap:ablation:sessions:data}%
        }%
    \caption{Reconstruction results for varying number of sessions and submap sizes on HD-Map dataset (a) Increasing the number of sessions improves not just semantic F-Score, but also recall \emph{and} precision. (b) Semantic F-Score for varying submap sizes leading to different amount of transmitted data per mapped area.}%
    \label{fig:cloudmap:ablation:sessions}%
\end{figure}
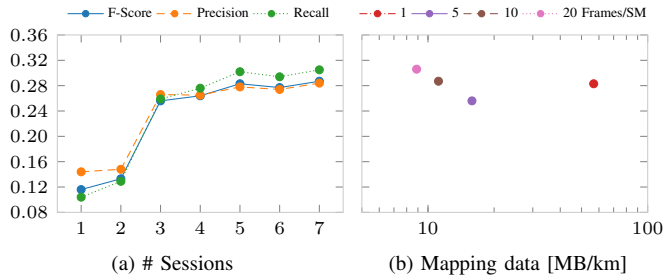

\subsubsection{Multi-Session Mapping}
In this section, we provide experimental results on our HD-Map dataset with varying number of total mapping sessions as well as with different submap sizes, which impacts the amount of data that needs to be collected from mapping agents.
We ran our reconstruction for each number of sessions $N \in [1, 7]$ with 3 randomly sampled sets each from the entire dataset for both forward and backward direction.
In \refFigure{fig:cloudmap:ablation:sessions:count} we can observe that with an increasing number of mapping sessions the reconstruction quality rises as well, supporting our motivation for multi-session mapping. Notably both recall \emph{and} precision increase with more sessions.
Further, we computed reconstructions using different submap sizes of $S = {1,5,10,20}$ keyframes each. Larger submaps are more compact as the relative overlap between consecutive maps is smaller, resulting in less data transmission and storage, while shorter submaps may be less prone to accumulating drift and fusion errors.
We can observe in \refFigure{fig:cloudmap:ablation:sessions:data} that larger submaps perform just as good or better than smaller submaps, while requiring significantly less data per kilometer of mapped road. However, the data overhead advantage becomes less relevant for sizes $S>10$. Interestingly, larger submaps lead to slightly superior reconstruction results compared to smaller ones, which we attribute to better alignment due to larger overlap between distinct submaps.

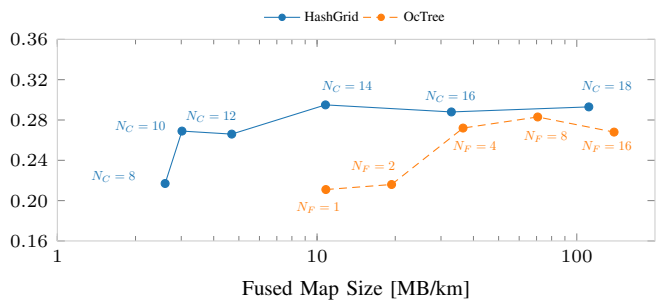
\begin{figure}[t!]
    \pgfplotstableread{figures/eval/data/grid/ngp.dat}\ablationgridngp%
\pgfplotstableread{figures/eval/data/grid/octree.dat}\ablationgridoctree%
\tikzset{PlotLabel/.style={scale=0.5,anchor=base west}}%
\begin{tikzpicture}[inner frame sep=0, trim axis right] %
    \begin{axis}[%
        PlotStyleHalfColumn,%
        width=0.92\columnwidth,%
        height=0.5\columnwidth/\goldenratio,%
        xlabel=Fused Map Size \lbrack MB/km\rbrack,%
        xmin=1,%
        xmax=200,%
        xmode=log,%
        log ticks with fixed point,%
        trim axis left,%
        ymax=0.36,%
        ymin=0.16,%
        ytick distance=0.04,%
        y tick label style={%
        /pgf/number format/.cd,%
            fixed,%
            fixed zerofill,%
            precision=2,%
        /tikz/.cd%
        },%
        legend columns=3,%
        legend style={%
            at={(0.5,1.03)},%
            anchor=south,%
            draw=none,%
            nodes={scale=0.75, transform shape}%
        }%
    ]%
    \addplot+ table[x=map_size_mb_km,y=fscore_mean]{\ablationgridngp};%
    \addplot+ table[x=map_size_mb_km,y=fscore_mean]{\ablationgridoctree};%
        \draw (axis cs:1.3,0.22)    node[PlotLabel,cblue]{$N_C=8$};%
        \draw (axis cs:1.6,0.27)    node[PlotLabel,cblue]{$N_C=10$};%
        \draw (axis cs:3,0.28)    node[PlotLabel,cblue]{$N_C=12$};%
        \draw (axis cs:10,0.31)    node[PlotLabel,cblue]{$N_C=14$};%
        \draw (axis cs:25,0.30)    node[PlotLabel,cblue]{$N_C=16$};%
        \draw (axis cs:100,0.31)    node[PlotLabel,cblue]{$N_C=18$};%
        \draw (axis cs:8,0.19)    node[PlotLabel,corange]{$N_F=1$};%
        \draw (axis cs:13,0.23)    node[PlotLabel,corange]{$N_F=2$};%
        \draw (axis cs:32,0.25)    node[PlotLabel,corange]{$N_F=4$};%
        \draw (axis cs:60,0.26)    node[PlotLabel,corange]{$N_F=8$};%
        \draw (axis cs:100,0.25)    node[PlotLabel,corange]{$N_F=16$};%
    \legend{HashGrid, OcTree}%
    \end{axis}%
\end{tikzpicture}%
%
    \caption{(a) Comparison of semantic F-Score for different HashGrid codebook sizes $N_C$ and OcTree feature sizes $N_F$ resulting in different map sizes (neural field weights).}%
    \label{fig:cloudmap:ablation:grid}%
\end{figure}

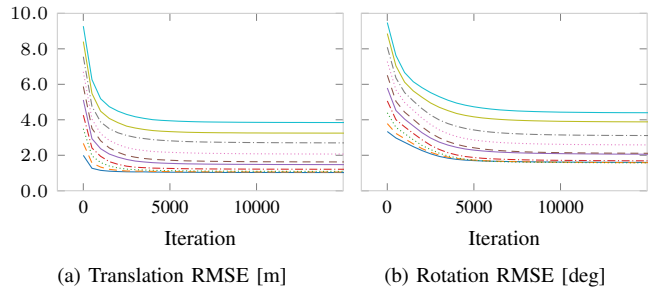
\begin{figure}[t!]
    \subfloat[Translation RMSE \lbrack m\rbrack]{%
        \pgfplotstableread{figures/eval/data/ablation_noise.dat}\ablationnoise%
\tikzset{PlotLabel/.style={scale=0.5,anchor=base west}}%
\begin{tikzpicture}[inner frame sep=0, trim axis right] %
    \begin{axis}[
        PlotStyleHalfColumn,
        cycle list name=tab10flat,
        trim axis left,
        width=0.44\columnwidth,
        height=0.44\columnwidth/\goldenratio,
        xlabel=Iteration,
        xtick={0,5000,10000},
        xticklabels={0,5000,10000},
        scaled x ticks=false,
        xmax=15000,
        ymax=10.0,
        ymin=0.,
        ytick distance=2,
        mark=none,
        y tick label style={
        /pgf/number format/.cd,
            fixed,
            fixed zerofill,
            precision=1,
        /tikz/.cd
        },
    ]
    \addplot+ table[x=iter,y=1v4-0_aligned_translation_part_rmse]{\ablationnoise};
    \addplot+ table[x=iter,y=1.5v4-0_aligned_translation_part_rmse]{\ablationnoise};
    \addplot+ table[x=iter,y=2v4-0_aligned_translation_part_rmse]{\ablationnoise};
    \addplot+ table[x=iter,y=2.5v4-0_aligned_translation_part_rmse]{\ablationnoise};
    \addplot+ table[x=iter,y=3v4-0_aligned_translation_part_rmse]{\ablationnoise};
    \addplot+ table[x=iter,y=3.5v4-0_aligned_translation_part_rmse]{\ablationnoise};
    \addplot+ table[x=iter,y=4v4-0_aligned_translation_part_rmse]{\ablationnoise};
    \addplot+ table[x=iter,y=4.5v4-0_aligned_translation_part_rmse]{\ablationnoise};
    \addplot+ table[x=iter,y=5v4-0_aligned_translation_part_rmse]{\ablationnoise};
    \addplot+ table[x=iter,y=5.5v4-0_aligned_translation_part_rmse]{\ablationnoise};
    \end{axis}%
\end{tikzpicture}%
%
        \label{fig:cloudmap:ablation:noise:trans}%
    }%
    \subfloat[Rotation RMSE \lbrack deg\rbrack]{%
    \pgfplotstableread{figures/eval/data/ablation_noise.dat}\ablationnoiserot%
\tikzset{PlotLabel/.style={scale=0.5,anchor=base west}}%
\begin{tikzpicture}[inner frame sep=0, trim axis right] %
    \begin{axis}[
        PlotStyleHalfColumn,
        cycle list name=tab10flat,
        trim axis right,
        width=0.44\columnwidth,
        height=0.44\columnwidth/\goldenratio,
        xlabel=Iteration,
        xtick={0,5000,10000},
        xticklabels={0,5000,10000},
        scaled x ticks=false,
        xmax=15000,
        ymax=10.0,
        ymin=0.0,
        ytick distance=2,
        yticklabels={},
    ]
    \addplot+ table[x=iter,y=1v4-0_aligned_rotation_angle_in_degrees_rmse]{\ablationnoiserot};
    \addplot+ table[x=iter,y=1.5v4-0_aligned_rotation_angle_in_degrees_rmse]{\ablationnoiserot};
    \addplot+ table[x=iter,y=2v4-0_aligned_rotation_angle_in_degrees_rmse]{\ablationnoiserot};
    \addplot+ table[x=iter,y=2.5v4-0_aligned_rotation_angle_in_degrees_rmse]{\ablationnoiserot};
    \addplot+ table[x=iter,y=3v4-0_aligned_rotation_angle_in_degrees_rmse]{\ablationnoiserot};
    \addplot+ table[x=iter,y=3.5v4-0_aligned_rotation_angle_in_degrees_rmse]{\ablationnoiserot};
    \addplot+ table[x=iter,y=4v4-0_aligned_rotation_angle_in_degrees_rmse]{\ablationnoiserot};
    \addplot+ table[x=iter,y=4.5v4-0_aligned_rotation_angle_in_degrees_rmse]{\ablationnoiserot};
    \addplot+ table[x=iter,y=5v4-0_aligned_rotation_angle_in_degrees_rmse]{\ablationnoiserot};
    \addplot+ table[x=iter,y=5.5v4-0_aligned_rotation_angle_in_degrees_rmse]{\ablationnoiserot};
    \end{axis}%
\end{tikzpicture}%
%
        \label{fig:cloudmap:ablation:noise:rot}%
        }%
    \caption{Comparison of (a) translational and (b) rotational alignment RMSE over iterations for different initial offsets.}%
    \label{fig:cloudmap:ablation:noise}%
\end{figure}

\subsubsection{Neural Field Grid}
In \refFigure{fig:cloudmap:ablation:grid} we compare the effect of different HashGrid and OcTree parameterizations on the reconstruction quality and the size of the neural field in terms of storage. We can clearly see that HashGrids are much more efficient in our setup compared to OcTrees, requiring around 5$\times$ less storage to achieve the same reconstruction quality.

\subsubsection{Alignment Robustness}
Since our joint reconstruction and pose-alignment is optimized towards a local minimum, reasonable initialization of poses is crucial to achieve good alignment.
We investigate the effect of inaccurate initial poses for various amounts of disturbance on KITTI GPS poses in \refFigure{fig:cloudmap:ablation:noise}.
We apply varying amount of artificial noise $e_\text{trans} \sim \normal(0, \sigma_\text{trans})$ and $e_\text{rot} \sim \normal(0, \sigma_\text{rot})$ for translational and rotational components with $\sigma_\text{trans} \in [1, 5.5]m$, $\sigma_\text{rot} \in [1, 5.5]^\circ$.
While low to medium levels of disturbance all convergence well to similar low final alignment errors, we find that the alignment result degrades well for increasing initial errors. Even for extreme levels of initial disturbance, there is a significant reduction in pose error, proving the robustness of joint reconstruction and alignment using neural field fusion.

\subsubsection{Confidence}
\begin{figure}[t!]
    \subfloat[Semantic / Confidence]{%
        \includegraphics[width=0.45\columnwidth,height=0.4\columnwidth]{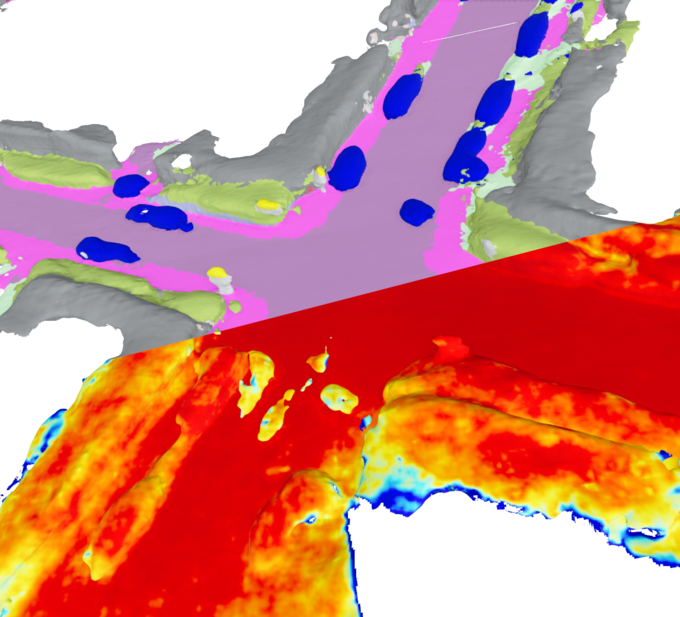}%
        \label{fig:cloudmap:ablation:confidence:visu:sem}%
    }%
    \subfloat[Confidence Threshold]{%
    \pgfplotstableread{figures/eval/data/ablation_weight.dat}\ablationweightdata
\pgfplotstableread{figures/eval/data/ablation_weight.dat}\ablationweightdatahd

\begin{tikzpicture}[inner frame sep=0, trim axis right] %
    \begin{axis}[
        PlotStyleHalfColumn,
        width=0.44\columnwidth,
        height=0.44\columnwidth/\goldenratio,
        xtick={0.1,0.3,0.5,0.7,0.9},
        ymax=0.36,
        ymin=0.16,
        ytick distance=0.04,
        y tick label style={
        /pgf/number format/.cd,
            fixed,
            fixed zerofill,
            precision=2,
        /tikz/.cd
        },
        legend columns=3,
        legend style={
            at={(0.5,1.03)},
            anchor=south,
            draw=none,%
            nodes={scale=0.75, transform shape}
        }
    ]
        \addplot+  table[x=weight,y=fscore_mean]{\ablationweightdata};
        \addplot+  table[x=weight,y=precision_mean]{\ablationweightdata};
        \addplot+  table[x=weight,y=recall_mean]{\ablationweightdata};
        \legend{F-Score, Precision, Recall}
    \end{axis}
\end{tikzpicture}%
    \label{fig:cloudmap:ablation:confidence:chart}%
    }%
    \caption{(a) Visualization of semantic (top left) and confidence score prediction (bottom right) with warmer colors denoting higher confidence. (b) Semantic scores for varying confidence thresholds $c_\text{th}$.}%
    \label{fig:cloudmap:ablation:confidence:visu}%
\end{figure}
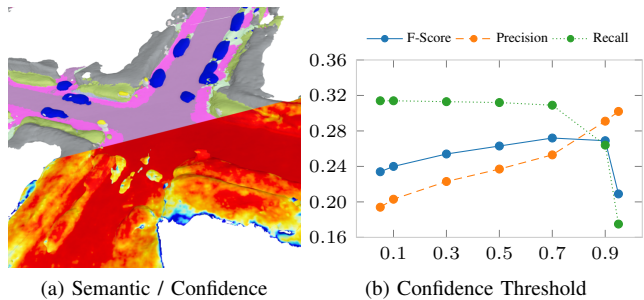

We show a visualization of our confidence prediction in \refFigure{fig:cloudmap:ablation:confidence:visu} combined with semantic prediction.
We observe that the main road areas carry a high confidence, visualized in red, while areas that are rarely observed or occluded have a low-confidence prediction shown in blue.
\refFigure{fig:cloudmap:ablation:confidence:chart} depicts the reconstruction F-score, precision and recall for various thresholds $c_\text{tb} \in [0, 1]$. We note that higher thresholds increase precision (accuracy) while decreasing recall (completeness) and find $c_\text{th}=0.7$ as balance that achieves good results across our datasets.

\section{Conclusion}
In this paper, we presented a novel multi-session mapping approach for crowd-sourced map learning.
Our system fuses submaps reconstructed by individual vehicles into a coherent dense semantic map that exceeds the fidelity of existing approaches for large-scale map learning.
Using meshes for transmitting map data to the mapping server enables lightweight communication and re-uses locally fused information from arbitrary sensors and reconstruction methods.
In turn, our neural field fusion facilitates large-scale joint alignment and 3D reconstruction without the need for dedicated localization landmarks.
In our experiments, we demonstrated the quality and robustness of the fused maps and showed the practicality and effectiveness of our approach for high-fidelity map learning.
We see neural implicit representations as exchange format from vehicle to server as well as change detection as promising directions for future research.
\bibliographystyle{bibliography/IEEEtran}%
\bibliography{references}%
\end{document}